\documentclass[10pt,twocolumn,letterpaper]{article}

\usepackage{wacv}
\usepackage{times}
\usepackage{epsfig}
\usepackage{graphicx}
\usepackage{amsmath}
\usepackage{amssymb}
\usepackage{placeins}
\usepackage{soul}

\usepackage[inline]{enumitem}
\usepackage{multirow}
\usepackage{booktabs}
\usepackage{ctable}
\usepackage{textcomp}

\usepackage[pagebackref=true,breaklinks=true,letterpaper=true,colorlinks,bookmarks=false]{hyperref}

\wacvfinalcopy 

\def\wacvPaperID{****} 

\ifwacvfinal\fi
\begin{document}

\title{Dense 3D Point Cloud Reconstruction Using a Deep Pyramid Network}

\author{Priyanka Mandikal \hspace{2cm} R. Venkatesh Babu \\
Video Analytics Lab, CDS, Indian Institute of Science, Bangalore, India\\
{\tt\small priyanka.mandikal@gmail.com, venky@iisc.ac.in}
}

\maketitle
\ifwacvfinal\thispagestyle{empty}\fi

\begin{abstract}
Reconstructing a high-resolution 3D model of an object is a challenging task in computer vision. Designing scalable and light-weight architectures is crucial while addressing this problem. Existing point-cloud based reconstruction approaches directly predict the entire point cloud in a single stage. Although this technique can handle low-resolution point clouds, it is not a viable solution for generating dense, high-resolution outputs. In this work, we introduce DensePCR, a deep pyramidal network for point cloud reconstruction that hierarchically predicts point clouds of increasing resolution. Towards this end, we propose an architecture that first predicts a low-resolution point cloud, and then hierarchically increases the resolution by aggregating local and global point features to deform a grid. Our method generates point clouds that are accurate, uniform and dense. Through extensive quantitative and qualitative evaluation on synthetic and real datasets, we demonstrate that DensePCR outperforms the existing state-of-the-art point cloud reconstruction works, while also providing a light-weight and scalable architecture for predicting high-resolution outputs.
\end{abstract}

\section{Introduction}
\label{sec:intro}

\begin{figure}[!hb]
\centering
\begin{center}
\includegraphics[width=\linewidth]{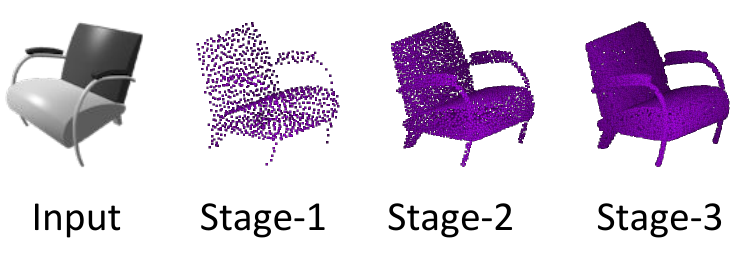}
\end{center}
\caption{Hierarchical reconstruction of a 3D model in stages.}
\label{fig:introduction}
\end{figure}

We inhabit a world of illuminated physical objects having diverse shapes, sizes and textures. The human visual system is capable of processing the retinal image of an object to extract the underlying 3D structure. Our 3D perception capabilities go beyond mere reconstruction of the overall shape. We are highly adept at capturing finer details present on the object surface. This ability to effectively capture accurate 3D characteristics is vital to our understanding of the physical world and manipulation of objects within it.

Machines would greatly benefit from learning to perceive the 3D world as humans do. A number of tasks such as robot grasping, object manipulation, and navigation are inherently dependent upon an agent's ability to understand physical objects and scenes. Further, the ability to not only have an understanding of the overall shape, but to also model the two-dimensional surface manifold is a critical function in 3D perception. An ideal machine would have the capacity to hierarchically grow its understanding of a surface representation (Fig.~\ref{fig:introduction}).

With the recent advances in deep learning, the problem of 3D reconstruction has largely been tackled with the help of 3D-CNNs that generate a voxelized 3D occupancy grid. Unlike 2D images, where all the pixels add rich spatial and structural information, volumetric representations suffer from sparsity of information. The information needed to perceive the 3D structure is provided by surface voxels, while the voxels within the volume increase the representational and computational complexity with minimal addition in information. This representation is particularly inefficient to encode high resolution outputs since the surface to volume ratio diminishes further with increase in resolution. 3D CNNs are also compute heavy and add considerable overhead during training and inference. 

The drawbacks of volumetric representations are alleviated by 3D point clouds. Point clouds are a more efficient alternative, since the points are sampled on the surface of the object, and can effectively capture detailed surface information. An inherent challenge in processing a point cloud however is that it is unordered i.e. any permutation of the points doesn't alter the 3D structure that it represents. CNN architectures have traditionally been used for analyzing ordered data such as images, voxelized grids, etc. Extending the above for unordered representations such as point clouds has very recently been studied using architectures and loss formulations introduced in ~\cite{qi2017pointnet,qi2017pointnet++,fan2017point,su2018splatnet,li2018pointcnn}. However, existing point cloud based reconstruction works directly predict the full resolution point cloud in a single stage. This has a number of issues when it comes to predicting dense outputs. 
\begin{enumerate*}[label=\textbf{(\arabic*)}]
    \item First, there is a substantial increase in the number of network parameters, which makes it difficult to scale up such architectures.
    \item Second, loss formulations such as the Earth Mover Distance, which is frequently applied on point sets is computationally heavy and its application to dense outputs would add considerable memory and time overhead. As a result, dense predictions would fail to benefit from some of the  properties enforced by such loss formulations such as uniformity in points.
\end{enumerate*}

In this work, we seek to find answers to two important questions in the task of single-view reconstruction
\begin{enumerate*}[label=\textbf{(\arabic*)}]
    \item Given a two-dimensional image of an object, what is an efficient and scalable solution to infer a dense 3D point cloud reconstruction of it?
    \item How do we upscale light-weight sparse point cloud representations so as to approximate surfaces in higher resolutions?
\end{enumerate*}
To address the former issue, we use a deep pyramidal architecture that first predicts a low-density sparse point cloud using minimal parameters, and then hierarchically increases the resolution using a point processing framework. To achieve the latter, we propose a mechanism to deform small local grids around each point using neighbourhood terrain information as well as global shape properties. These learnt local grids now approximate a surface in the next stage. The benefits of this technique are two-fold. First, predicting a low resolution coarse point cloud enables us to utilize loss formulations that otherwise may be very intensive for dense point clouds. This leads to better quality predictions. Second, using a pyramidal approach on point sets drastically reduces the number of network parameters, a crucial operation for dense predictions.

In summary, our contributions in this work are as follows:
\begin{itemize}
    \item We propose a deep pyramidal network for point cloud reconstruction called DensePCR, that hierarchically predicts point clouds of increasing resolution. Intermediate point clouds are super-resolved by extracting global and local point features and conditioning them on a grid around each point, so as to approximate a surface in the higher resolution.
    \item We perform a network analysis of the proposed architecture to compare against existing approaches and demonstrate that densePCR has substantially fewer number of learnable parameters, an essential requirement for dense prediction networks. Specifically, our network gives rise to 3X reduction in the parameters as compared to the baseline point cloud reconstruction networks.
    \item We highlight the efficacy of this network in generating high quality predictions by evaluating on a large scale synthetic dataset, where we outperform the state-of-the-art approaches by a significant margin, despite having fewer parameters. 
    \item We evaluate DensePCR on real data and demonstrate the generalization ability of our approach, which significantly outperforms the state-of-art reconstruction methods. 
\end{itemize}
\section{Related Work}
\label{sec:related_work}

\noindent
\textbf{3D Reconstruction}
For a long time, the task of 3D reconstruction from single-view images had largely been tackled with the help of 3D CNNs. A number of works have revolved around generating voxelized output representations ~\cite{girdhar2016learning,wu2016learning,choy20163d,qi2016volumetric,kar2017learning,wu20153d,ulusoy2015towards}. Giridhar \etal ~\cite{girdhar2016learning} learnt a joint embedding of 3D voxel shapes and their corresponding 2D images. Choy \etal~\cite{choy20163d} trained a recurrent neural network to encode information from more than one input views. But voxel formats are computationally heavy and information-sparse, which lead to research on the octree data structure for representing 3D data~\cite{tatarchenko2017octree, Riegler2017CVPR, Riegler2017THREEDV, hspHane17, wang2017cnn}.

Recently, Fan \etal~\cite{fan2017point}, introduced techniques for generating unordered point clouds to obtain single-view 3D reconstruction results outperforming volumetric approaches~\cite{choy20163d}. While ~\cite{fan2017point} directly predict the 3D point cloud from 2D images, our approach stresses the importance of first predicting a low-resolution point cloud and super-resolving it to obtain a dense prediction. Groueix \textit{et al.}~\cite{groueix2018} represented a 3D shape as a collection of parametric surface elements and constructed a mesh from the predicted point cloud. Mandikal et al.~\cite{mandikal20183dlmnet} proposed a latent matching setup in a probabilistic framework to obtain diverse reconstructions. Other works utilize 2D supervision in the form of silhouettes and depth maps for 3D reconstruction~\cite{yan2016perspective, tulsiani2017multi, wu2017marrnet, zhu2017rethinking, gwak2017weakly,lin2018learning,kl2019capnet}. 
Concurrently with us, Yuan \textit{et al.}~\cite{yuan2018pcn} propose to deform grids for completing partial depth maps.
Apart from reconstruction, other 3D perception tasks have also been performed using point clouds~\cite{achlioptas2017representation,yu2018pu,mandikal20183dpsrnet}. \newline

\begin{figure*}[!ht]
\centering
\begin{center}
\includegraphics[width=\linewidth]{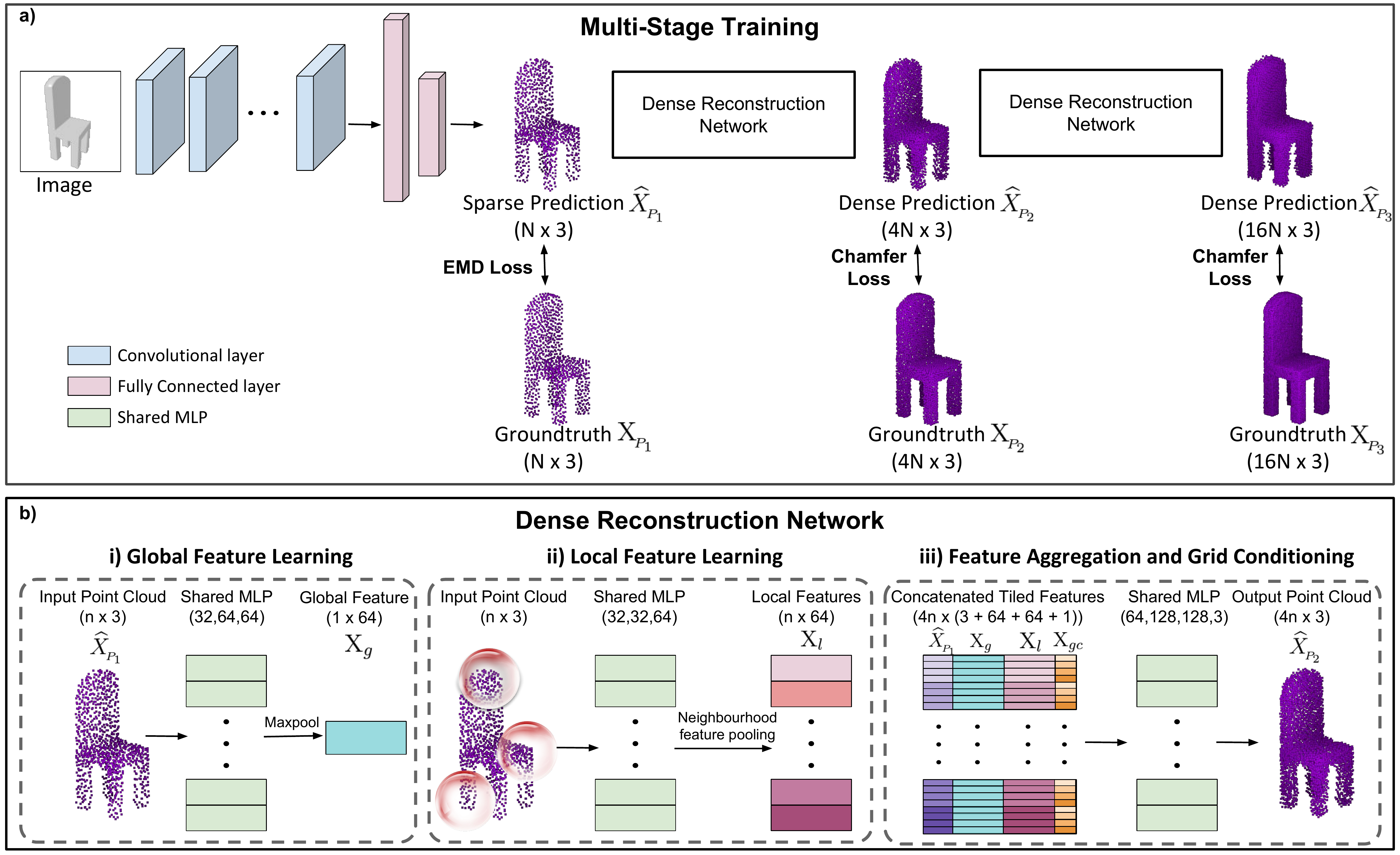}
\end{center}
\caption{Overview of DensePCR. \textbf{a)} The training pipeline consists of first predicting a low-density point cloud, and then hierarchically increasing the resolution. A multi-scale training strategy is utilized to enforce constraints on each of the intermediate outputs. \textbf{b)} Intermediate point cloud ($X_{\!_P}$) is super-resolved by extracting global ($X_g$) and local ($X_l$) point features and conditioning them on a coordinate grid ($X_c$) around each point to generate a dense prediction.}
\label{fig:proposed_arch}
\end{figure*}

\noindent
\textbf{Hierarchical Prediction}
The concept of Laplacian pyramid networks has been previously used in 2D vision tasks for hierarchical prediction. Denton \etal ~\cite{denton2015deep} proposed a generative adversarial network to generate realistic images based on a Laplacian pyramid framework (LAPGAN). Lai \etal~\cite{lai2018fast} extended the above by introducing a robust loss formulation and making architectural modifications for improving speed and accuracy. In the 3D vision domain, Hane \etal~\cite{hspHane17} proposed an octree-based method for hierarchical surface prediction. While the focus of~\cite{hspHane17} is on extending a volumetric representation into an octree-based one to enable surface prediction, we directly operate on points sampled on the object surface. 

Yu \etal~\cite{yu2018pu} proposed a network to upsample point clouds (PU-Net), which is more related to our work. However, the proposed DensePCR differs from PU-Net in two aspects. First, the objectives of the two models are different. PU-Net is a super-resolution model which is designed to upsample input point clouds which display high uniformity albeit with some resistance to small noise. On the contrary, the proposed DensePCR is a reconstruction model that predicts a particular 3D point cloud based on the given RGB image and then increases the resolution. It is important to note that the point clouds to be upsampled are far from ideal, unlike in the case of PU-Net. Second, PU-Net uses a repulsion loss to avoid point clustering and ensure that output points are well-separated. In contrast, we ensure that points don't cluster in higher resolutions by conditioning aggregated features on a local coordinate grid. This is akin to deforming small local grids around each point while upscaling. Furthermore, PU-Net uses the computationally expensive Earth Mover's Distance (EMD) loss at higher resolutions, while we use the more light-weight Chamfer distance metric.
\section{Approach}
\label{sec:approach}

Our training pipeline consists of a multi-stage training paradigm as outlined in Fig.~\ref{fig:proposed_arch}. The input RGB image is passed through an encoder-decoder network that outputs the reconstructed point cloud at a low resolution. This intermediate output is subsequently passed through a dense reconstruction network to obtain a high-resolution point cloud. This is done via aggregating global and local point features and conditioning them around a coordinate grid to approximate a surface around each point in the higher resolution. Point set distance based losses are enforced at every stage to ensure a coherency of intermediate outputs. Each of the components of our approach is described in detail below.

\subsection{Multi-Stage Training}
Our goal is to train a network to generate a dense 3D point cloud given a singl-view image. For this purpose, we train a deep pyramidal neural network in a hierarchical fashion (Fig.~\ref{fig:proposed_arch}a). An encoder-decoder network takes in an input RGB image $I$ and outputs a sparse 3D point cloud $\widehat{X}_{\!_P}$. Since a point cloud is an unordered form of representation, we need a loss formulation that is invariant to the relative ordering of input points. To enforce this, we can use one of two commonly-used loss functions on point sets - Chamfer Distance (CD) and Earth Mover's Distance (EMD). 
Chamfer distance between $\widehat{X}_{\!_P}$ and $X_{\!_P}$ is defined as:  
\begin{equation}
\label{eq:chamfer}
    \begin{split}
        d_{Chamfer}(\widehat{X}_{\!_P},X_{\!_P}) = \sum_{x\in \widehat{X}_{\!_P}}\min_{y\in X_{\!_P}}{||x-y||}^2_2 
        \\ + \sum_{y\in \widehat{X}_{\!_P}}\min_{x\in X_{\!_P}}{||x-y||}^2_2
    \end{split}
\end{equation}
Intuitively, this loss function is a nearest neighbour distance metric that computes the error in two directions. Every point in $\widehat{X}_{\!_P}$ is mapped to the nearest point in $X_{\!_P}$, and vice versa. Although this is a computationally light formalism, it has a critical drawback in that there is no mechanism to ensure uniformity of predicted points~\cite{achlioptas2017representation}. In other words, the optimization often leads to a minima where only a subset of points account for the overall shape, thereby resulting in the clustering of the other points. The EMD loss serves to alleviate this concern. EMD between two point sets $\widehat{X}_{\!_P}$ and $X_{\!_P}$ is given by:
\begin{equation}
\label{eq:emd}
        d_{EMD}(\widehat{X}_{\!_P},X_{\!_P})=\min_{\phi:\widehat{X}_{\!_P}\rightarrow X_{\!_P}}\sum_{x\in \widehat{X}_{\!_P}}||x-\phi(x)||_2
\end{equation}
where $\phi:\widehat{X}_{\!_P}\rightarrow X_{\!_P}$ is a bijection. 
Since it enforces a point-to-point mapping between the two sets, it ensures uniformity in point predictions. However, a major drawback of EMD is that it is a computationally intensive formulation, requiring considerable amount of time and memory for high-resolution point sets. Keeping this in mind, we design a training regime that can take advantage of both the losses, while avoiding the common pitfalls. The sparse point cloud $\widehat{X}_{\!_P}$ predicted at the first stage is optimized via the EMD loss. This ensures that the base point cloud to be upsampled is a uniform prediction. All subsequent point clouds are optimized via the Chamfer Distance. This design choice allows us to train the network so as to enjoy the benefits arising from both EMD as well as CD. The architecture of the Dense Reconstruction Network ensures that the upsampled points remain uniform and do not collapse onto one another.

\subsection{Dense Reconstruction Network}
We aim to effectively predict a dense point cloud given its sparse counterpart. Towards this end, we propose an architecture that processes the input point cloud and learns various features that aid in the upsampling process (Fig.~\ref{fig:proposed_arch}b). \vspace{0.5em}

\noindent
\textbf{Global Feature Learning}
An understanding of global shape properties is essential knowledge in order to reconstruct the input shape at a higher resolution. This is especially important when dealing with a diverse set of data, since point density and spacing are generally model-specific. To incorporate this knowledge into the learning framework, we make use of an MLP architecture similar to PointNet~\cite{qi2017pointnet}. It consists of a shared set of MLPs operating on every individual point in the input point cloud $X_{\!_P}$ of resolution $n\times3$ (Fig.~\ref{fig:proposed_arch}b-i). A point-wise max pooling operation is applied after the final layer output, to obtain the global feature vector $X_g$ having dimension $1\times n_g$, where $n_g$ is the number of filters in the final MLP. $X_g$ now contains salient properties defining the global shape characteristics. \vspace{0.5em}

\noindent
\textbf{Local Feature Learning}
The importance of local features in per-point prediction tasks such as segmentation and normal estimation has been established by~\cite{qi2017pointnet,qi2017pointnet++}. Local point features provide neighbourhood information that enable better capture of the local terrain. Such information is necessary in order to build models that can effectively upsample points to fill in gaps. Hence, we take the approach of~\cite{qi2017pointnet++} to extract local features around every input point. Specifically, we construct neighbourhood balls around every incoming point, and locally apply a sequence of MLPs on every neighbourhood. Neighbourhood features are then pooled to obtain the local feature for a point. Performing this operation on every point in the input point cloud $X_{\!_P}$ gives a set of local features $X_l$, having dimension $n\times n_l$, where $n_l$ is the number of filters in the final MLP. \vspace{0.5em}

\noindent
\textbf{Feature Aggregation and Coordinate Grid Conditioning}
Once we compute the global and local features, we require a mechanism to propagate these features to a higher resolution. For this purpose, we form a feature vector by concatenating the input points $X_{\!_P}$, tiled global feature $X_g$, and the local feature $X_l$ to obtain $[X_{\!_P}, X_g, X_l]$ of dimension $n\times(3+n_g+n_l)$. We then tile this feature vector by the upsampling factor to obtain a feature of dimension $4n\times(3+n_g+n_l)$. However, we require a mechanism to induce separability between features of the same point. We achieve this with the help of a 2D coordinate grid $X_{gc}$ of dimension $2\times2$ reshaped to $4\times1$, centered around each point (here \textit{gc} stand for 'grid conditioning'). The grid serves to condition the point features so that MLPs in the next layer can learn to propagate the point features. Intuitively, the network learns a set of parameters to deform the grid so as to approximate a surface around that point in the higher resolution. This grid is tiled for every input point to obtain a feature of dimension $4n\times1$. The final aggregated feature vector is given by $[X_{\!_P}, X_g, X_l, X_{gc}]$, having dimension $4n\times(3+n_g+n_l+1)$, which is operated upon by a set of shared MLPs to finally reconstruct the dense point cloud.

\subsection{Implementation Details}
The image encoder is a 2D CNN consisting of 16 convolutional layers and outputs a latent vector of dimension 512. Our network predicts point clouds at three stages of hierarchy - 1024, 4096 and 16384 points. The decoder which outputs the initial sparse point cloud consists of fully connected layers of dimension $[256,256,1024*3]$. Thereafter, the dense reconstruction network consists purely of  multi-layer perceptrons (MLPs). Specifically, we use MLPs of dimensions $[32,64,64]$ for the global feature learning module, $[32,32,64]$ for the local feature learning module, and $[32,64,64]$ in the feature aggregation module. The ball radius size for neighbourhood feature pooling in the local module is set to $0.1$ and $0.05$ for $1024$ and $4096$ points respectively. The grid used for feature conditioning is a $2\times2$ grid of size $0.2$ . We first pre-train the different stages of the network independently and later fine-tune the model end-to-end to obtain the final predictions. We notice that this training setup converges faster and is more stable than training the entire pyramid in an end-to-end manner from scratch. We use the Adam~\cite{kingma2014adam} optimizer with a learning rate of $0.00005$ and a minibatch size of 32. Network architectures for all components in our proposed framework are provided in the supplementary material.

\section{Experiments}
\label{sec:evaluation}

\subsection{Dataset}
We train all our networks on synthetic models from the ShapeNet~\cite{chang2015shapenet} dataset. We use the $80\%$-$20\%$ train/test split provided by~\cite{choy20163d} consisting of models from 13 different categories. We use the input images provided by~\cite{choy20163d}, where each model is pre-rendered from 24 different azimuth angles. We crop the images to $128\times 128$ resolution before passing it through our network. For generating the ground truth point clouds, we uniformly sample $N$, $4N$ and $16N$ points on the mesh surface using farthest point sampling, where $N=1024$.

\subsection{Evaluation Methodology}
We report both the Chamfer Distance (Eqn.~\ref{eq:chamfer}) as well as the Earth Mover's Distance (EMD, Eqn.~\ref{eq:emd}) computed on 16,384 points in all our evaluations. We use an approximate version of the EMD metric due to the high computational cost involved. For computing the metrics, we renormalize both the ground truth and predicted point clouds within a bounding box of length 1 unit. 

\subsection{Baselines}
We consider the PSGN network proposed by Fan \etal~\cite{fan2017point} as the baseline for the point cloud reconstruction task. Specifically, we consider two variants of PSGN - (1) PSGN-FC, and (2) PSGN-ConvFC in our evaluation. The two variants differ in their decoder architectures. PSGN-FC consists of a set of fully connected layers to directly predict the full resolution point cloud. PSGN-ConvFC utilizes both a deconvolution as well as fully connected network for point prediction. We use the same architecture as provided by the authors for training the networks, making necessary changes to modify the output to 16,384 points. Note that both variants of PSGN as well as our DensePCR network share the same architecture for the encoder, so as to be comparable. Our network augments the decoder with a dense reconstruction network, whereas PSGN predicts the output in a single stage. Detailed network architectures for the baselines and our network is provided in the supplementary.

\begin{figure*}[t]
\centering
\begin{center}
    \includegraphics[width=\linewidth]{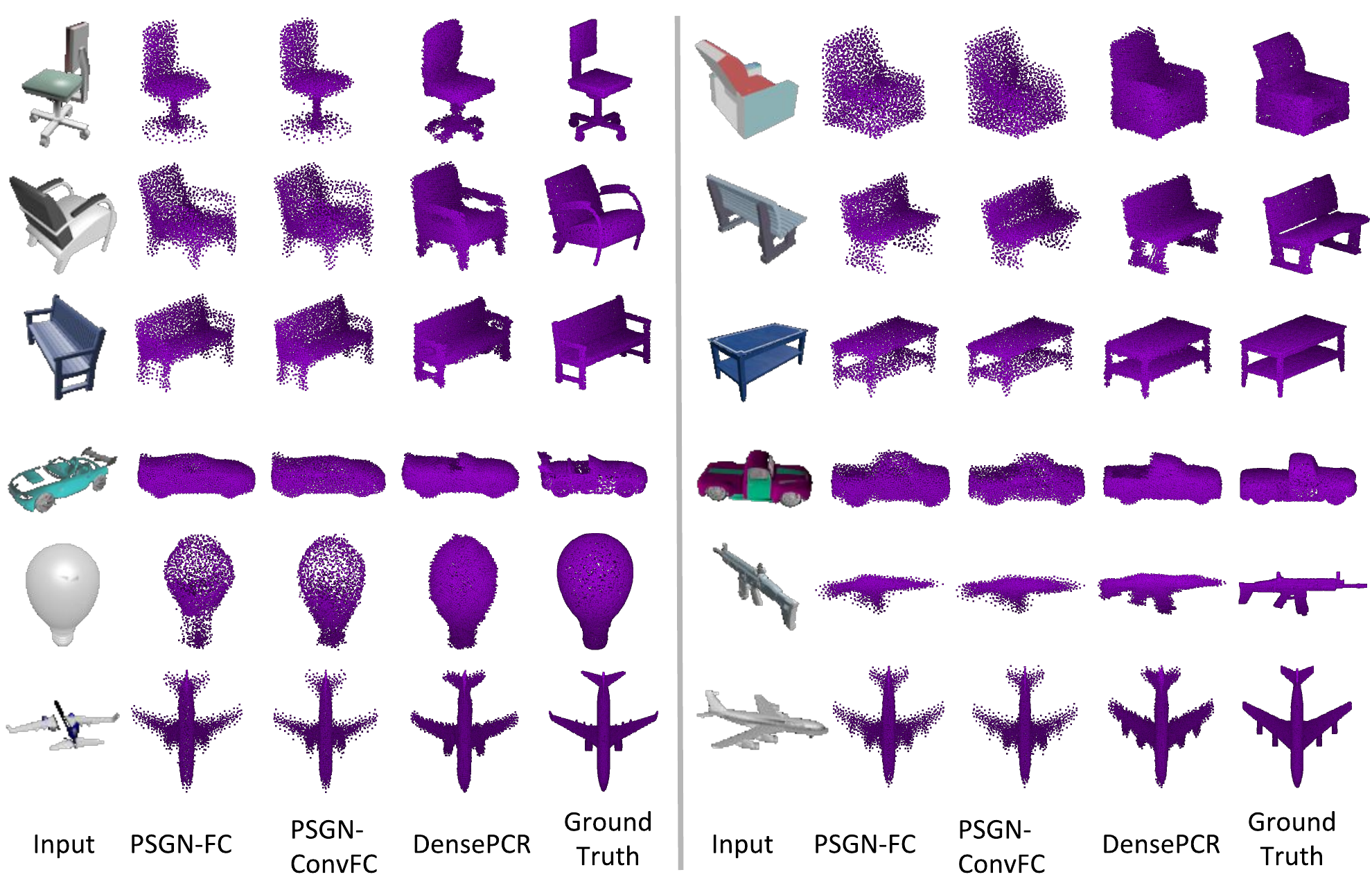}
\end{center}
\caption{Qualitative results on ShapeNet~\cite{chang2015shapenet}. Compared to both variants of PSGN~\cite{fan2017point},  we are better able to capture the overall shape and finer details present in the input image (legs of chairs, wings and tail of airplanes, trigger of rifles). Note that both the variants of PSGN have heavily clustered regions in many of the predictions while our reconstructions are more uniformly distributed.}
\label{fig:shapenet_comparison}
\end{figure*}

\subsection{Evaluation on ShapeNet}
 We compare our DensePCR model with the two variants of PSGN~\cite{fan2017point} on the synthetic ShapeNet dataset~\cite{chang2015shapenet}. Since~\cite{fan2017point} establishes that the point cloud based approach significantly outperforms the state-of-art voxel based approaches~\cite{choy20163d}, we do not show any comparison against them. Table~\ref{tab:sota_shapenet} shows the comparison between PSGN~\cite{fan2017point} and our method on the validation split provided by~\cite{choy20163d}. We outperform both variants of PSGN in 5 out of 13 categories in Chamfer and all 13 categories in the EMD metric, while also having lower overall mean scores. The gain in the EMD metric can be attributed to our multi-stage training setup, which enables us to use the EMD loss at the coarse stage and the light-weight Chamfer loss along with grid-conditioning at the fine stage, thus enforcing uniformity as well as maintaining it at latter stages. It is worth noting that we achieve state-of-the-art performance in both metrics despite the fact that our network has one third the number of trainable parameters in comparison to PSGN. A lower EMD score also correlates with better visual quality and encourages points to lie closer to the surface~\cite{achlioptas2017representation,yu2018pu}. Note that the numbers for EMD are much higher than that for Chamfer, since EMD places a high penalty on clusters as it is a point-to-point distance matching metric. Qualitative comparison is shown in Fig.~\ref{fig:shapenet_comparison}. Compared to PSGN, we are better able to capture the overall shape and finer details present in the input image (legs of chairs, wings and tail of airplanes, trigger of rifles). Note that both the variants of PSGN have heavily clustered regions in many of the predictions while our reconstructions are more uniformly distributed. 
 
 The uniformity in our predictions can be attributed to the loss formulations we adopt at different stages of the network. Specifically, enforcing the EMD loss at the coarse stage forces the predictions at the first stage to be uniform. Since subsequent stages are trained to upsample the coarse output, the predictions tend to be uniform despite using the Chamfer distance as the loss. Hence, a careful selection of losses at various stages enables us to train a model with lower computational complexity that achives better performance.

\begin{table*}[]
\centering
\begin{center}
\begin{tabular}{|c|ccc|ccc|}
\hline
\multirow{2}{*}{Category}      & \multicolumn{3}{c|}{Chamfer} & \multicolumn{3}{c|}{EMD} \\ \cline{2-7} 
& PSGN-FC~\cite{fan2017point}   & PSGN-ConvFC~\cite{fan2017point}   & DensePCR  & PSGN-FC~\cite{fan2017point}   & PSGN-ConvFC~\cite{fan2017point}   & DensePCR \\ \hline\hline
airplane  & \textbf{4.98}  & 5.11  & 5.25  & 3.84 & 3.87 & \textbf{1.64} \\
bench     & \textbf{6.91}  & 7.08  & 6.99  & 4.34 & 4.53 & \textbf{1.35} \\
cabinet   & 10.09 & 10.23 & \textbf{8.70}  & 6.42 & 6.37 & \textbf{1.78} \\
car       & \textbf{4.84}  & 5.21  & 5.04  & 3.74 & 4.75 & \textbf{1.24} \\
chair     & \textbf{9.60}  & 9.75  & 9.83  & 5.94 & 5.94 & \textbf{2.01} \\
lamp      & \textbf{13.38} & 13.57 & 14.24 & 6.87 & 6.90 & \textbf{2.64} \\
monitor   & \textbf{11.54} & 12.01 & 11.91 & 6.27 & 6.10 & \textbf{2.08} \\
rifle     & 4.60  & \textbf{4.52}  & 4.59  & 3.86 & 3.86 & \textbf{1.82} \\
sofa      & 8.66  & 8.93  & \textbf{8.55}  & 4.88 & 4.89 & \textbf{1.68} \\
speaker   & 14.79 & 15.78 & \textbf{13.27} & 6.97 & 6.74 & \textbf{2.24} \\
table     & 9.83  & 9.87  & \textbf{9.20}  & 6.22 & 6.15 & \textbf{1.83} \\
telephone & 7.14  & 7.76  & \textbf{7.07}  & 5.26 & 5.28 & \textbf{1.61} \\
vessel    & \textbf{7.23}  & 7.25  & 7.54  & 4.27 & 4.23 & \textbf{1.93} \\ \hline
\textbf{mean} & 8.74  & 9.01  & \textbf{8.63}  & 5.30 & 5.35 & \textbf{1.83} \\ \hline
\end{tabular}
\end{center}
\caption{Single view reconstruction results on ShapeNet~\cite{chang2015shapenet}. Chamfer metrics are scaled by 100. EMD metrics are scaled by 10. The proposed method is comparable to or better than PSGN~\cite{fan2017point} on most categories in the Chamfer metric, and significantly better in all categories on the EMD metric.}
\label{tab:sota_shapenet}
\end{table*}

\begin{figure*}[t]
\centering
\begin{center}
    \includegraphics[width=\linewidth]{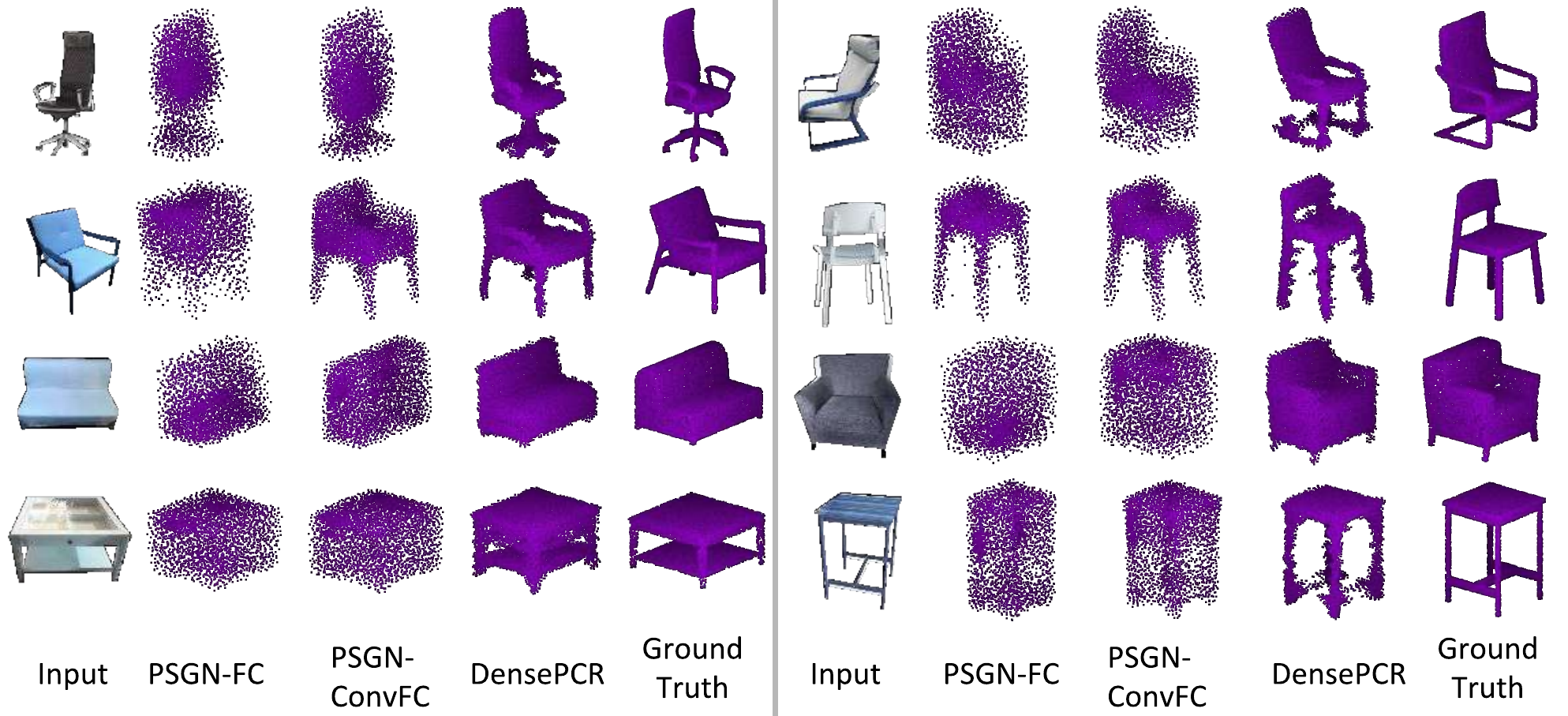}
\end{center}
\caption{Reconstructions on real-world Pix3D (chair, sofa, table). Surprisingly, both variants of PSGN have very poor generalizability, predicting highly incoherent shapes that often do not correspond to the input image (especially in chairs and sofas). On the other hand, DensePCR has very good generalization capability and predicts shapes that display high correspondence with the input image, despite the input space coming from a different distribution. Note that all three networks are trained on the same ShapeNet training set and tested on Pix3D.}
\label{fig:pix3d_reconstructions}
\end{figure*}

\begin{table*}[]
\centering
\begin{center}
\begin{tabular}{|c|ccc|ccc|}
\hline
\multirow{2}{*}{Category}      & \multicolumn{3}{c|}{Chamfer} & \multicolumn{3}{c|}{EMD} \\ \cline{2-7} 
& PSGN-FC~\cite{fan2017point}   & PSGN-ConvFC~\cite{fan2017point}   & DensePCR  & PSGN-FC~\cite{fan2017point}   & PSGN-ConvFC~\cite{fan2017point}   & DensePCR \\ \hline\hline
chair  & 8.39  & 8.16  & \textbf{7.52}  & 2.95 & 2.88 & \textbf{1.30} \\
sofa     & 9.72  & 9.50  & \textbf{6.94}  & 2.88 & 3.03 & \textbf{1.10} \\
table    & 12.09 & \textbf{11.02} & 11.21  & 3.65 & 3.51 & \textbf{1.63} \\ \hline
\textbf{mean} & 10.06 & 9.56  & \textbf{8.56}  & 3.16 & 3.14 & \textbf{1.34} \\ \hline
\end{tabular}
\end{center}
\caption{Single view reconstruction results on the real world Pix3D dataset~\cite{pix3d}. Chamfer metrics are scaled by 100, EMD metrics are scaled by 10.}
\label{tab:sota_pix3d}
\end{table*}

\section{Evaluation on Real-World Pix3D}
We evaluate the performance of our method on the Pix3D dataset~\cite{pix3d} to test the generalizability of our approach on real-world datasets. The dataset consists of a large collection of real images and their corresponding metadata such as masks, ground truth CAD models and pose. We evaluate our trained model on categories that co-occur in the ShapeNet training set and exclude images having occlusion and truncation from the test set, as is done in the original paper~\cite{pix3d}. We crop the images to center-position the object of interest and mask the background using the provided mask information. Table~\ref{tab:sota_pix3d} contains the results of this evaluation. Evidently, we outperform both variants of PSGN~\cite{fan2017point} by a large margin in both Chamfer as well as EMD metrics, demonstrating the effectiveness of our approach on real data. Fig.~\ref{fig:pix3d_reconstructions} shows sample reconstructions on this dataset. Surprisingly, both variants of PSGN have very poor generalizability, predicting highly incoherent shapes that often do not correspond to the input image (especially in chairs and sofas). On the other hand, DensePCR has very good generalization capability and predicts shapes that display high correspondence with the input image, in spite of the input images being from a different distribution. Note that all three networks are trained on the same ShapeNet training set and tested on Pix3D.

\section{Discussion}
\label{sec:disucssion}

\subsection{Network Parameters}
All the networks used in performing the experiments consist of the same architecture for the image encoder (~8.1M parameters) and vary with respect to the decoder. The decoder in PSGN-FC consists of only fully connected layers to directly predict 16,384 points, thus having a sizeable number of parameters at ~17.1M. The PSGN-ConvFC decoder consists of both deconvolution layers as well as fully connected layers and has ~13M parameters. Our network consists of fully connected decoder to predict the initial sparse prediction, followed by extremeley light-weight MLPs thereafter. This enables us to train our network with as little as ~5.2M parameters, almost a 3X reduction compared to both the PSGN variants. This makes our network highly apt for scaling up to very dense reconstructions with minimal addition in parameters. It is worth noting that the majority of parameters in our setup are from the initial base point cloud prediction network. The dense reconstruction network has very few parameters (0.043M parameters per every stage of hierarchy). Hence, as we scale up, there will be negligible addition of parameters, making it highly efficient for dense prediction. Detailed network architecture for each of the models is provided in the supplementary.

\subsection{Intermediate Outputs}
\begin{figure}
\centering
\begin{center}
    \includegraphics[width=0.95\linewidth]{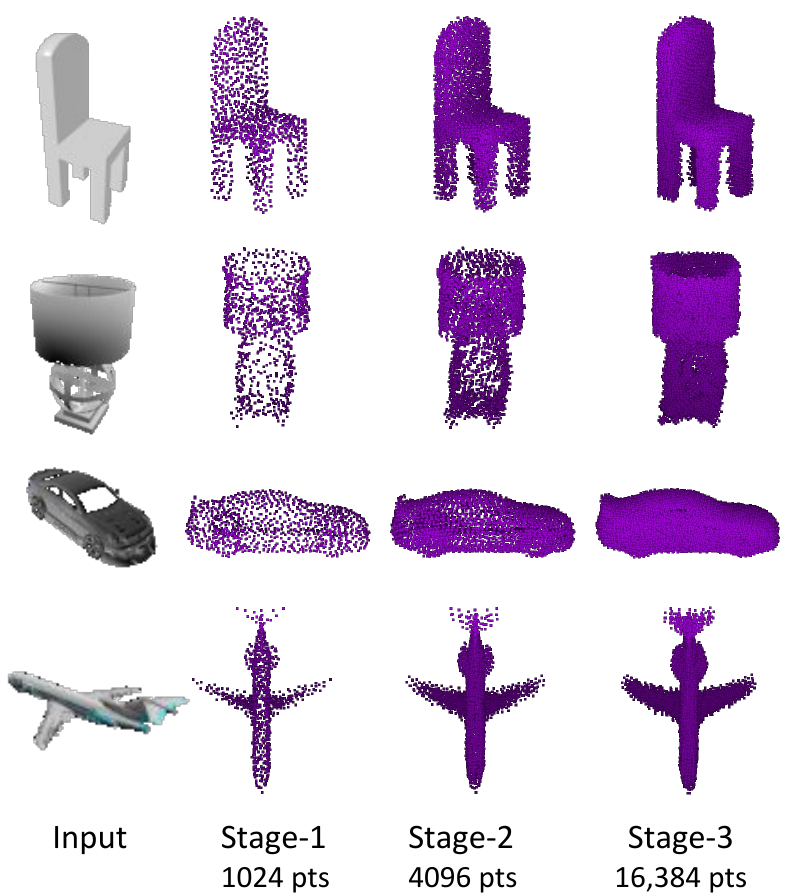}
\end{center}
\caption{Network predictions at different stages of the hierarchy. We observe that the dense reconstruction network is able to effectively scale up the point cloud resolution, while also retaining the uniformity of points present in the initial prediction.}
\label{fig:hierarchy_stages}
\end{figure}

We analyze the network predictions at the intermediate stages of the pyramidal network in Fig.~\ref{fig:hierarchy_stages}. We observe that the initial sparse prediction has good correspondence with the input image, while also having uniformity in point prediction due to the constraints enforced by the EMD loss (Eqn.~\ref{eq:emd}). The dense reconstruction network effectively upsamples this prediction to obtain dense point clouds that maintain the uniformity despite being trained with the Chamfer loss (Eqn.~\ref{eq:chamfer}). This can be attributed to the grid conditioning mechanism that prevents points from collapsing to a single region in space.

\subsection{Failure Cases}
\begin{figure}
\centering
\begin{center}
    \includegraphics[width=0.8\linewidth]{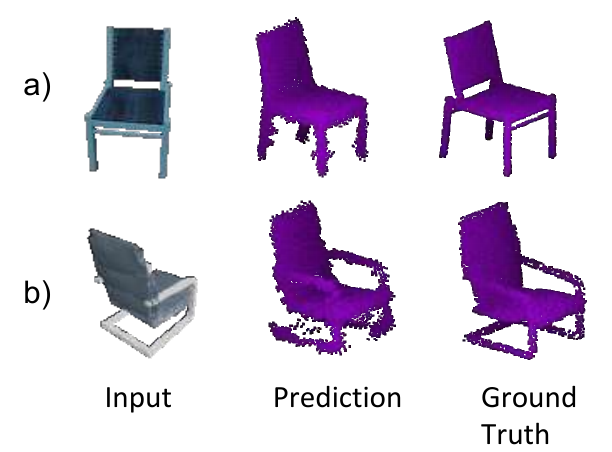}
\end{center}
\caption{Failure cases of our method. a) Certain predictions introduce artifacts such as a small clusters at the dense stage. b) Absence or distortion of finer details is observed in some models.}
\label{fig:failure_cases}
\end{figure}

We present some failure cases of our method in Fig.~\ref{fig:failure_cases}. We notice that certain predictions have artifacts consisting of small cluster of points around some regions. This is due to outlier points in the sparse point cloud, which get aggregated in the dense reconstruction. Utilizing a cascade of multiple neighbourhood features centered around each point might help in alleviating some of these issues, since it would be able to handle a higher variance in point cloud density. We also observe that certain predictions distort finer details present in the input image. Propagating the image feature at the dense reconstruction stage might aid in better capturing such details.

\section{Conclusion}
\label{sec:conclusion}
In this paper, we proposed a deep pyramidal network for hierarchical prediction of dense point clouds from a single RGB image. We highlighted the memory efficiency and scalability of this network in predicting high-resolution outputs. 
We presented a technique to approximate a surface around points by deforming local grids via aggregating local and global point features. Quantitative and qualitative evaluation on the single-image 3D reconstruction task on a synthetic dataset showed that our method generates high quality dense point clouds, while also providing a light-weight and scalable framework for point cloud prediction. Furthermore, evaluation on a real-world dataset displayed the superior generalization capability of our approach to new and unseen datasets.

{\small
\bibliographystyle{ieee}
\bibliography{main}

\begin{thebibliography}{10}\itemsep=-1pt

\bibitem{achlioptas2017representation}
P.~Achlioptas, O.~Diamanti, I.~Mitliagkas, and L.~Guibas.
\newblock Representation learning and adversarial generation of {3D} point
  clouds.
\newblock In {\em ICML}, 2018.

\bibitem{chang2015shapenet}
A.~X. Chang, T.~Funkhouser, L.~Guibas, P.~Hanrahan, Q.~Huang, Z.~Li,
  S.~Savarese, M.~Savva, S.~Song, H.~Su, et~al.
\newblock Shapenet: An information-rich {3D} model repository.
\newblock {\em arXiv preprint arXiv:1512.03012}, 2015.

\bibitem{choy20163d}
C.~B. Choy, D.~Xu, J.~Gwak, K.~Chen, and S.~Savarese.
\newblock {3D-r2n2}: A unified approach for single and multi-view {3D} object
  reconstruction.
\newblock In {\em ECCV}, 2016.

\bibitem{denton2015deep}
E.~L. Denton, S.~Chintala, R.~Fergus, et~al.
\newblock Deep generative image models using a￼ laplacian pyramid of
  adversarial networks.
\newblock In {\em NeurIPS}, 2015.

\bibitem{fan2017point}
H.~Fan, H.~Su, and L.~Guibas.
\newblock A point set generation network for {3D} object reconstruction from a
  single image.
\newblock In {\em CVPR}, volume~38, 2017.

\bibitem{girdhar2016learning}
R.~Girdhar, D.~F. Fouhey, M.~Rodriguez, and A.~Gupta.
\newblock Learning a predictable and generative vector representation for
  objects.
\newblock In {\em European Conference on Computer Vision}, pages 484--499.
  Springer, 2016.

\bibitem{groueix2018}
T.~Groueix, M.~Fisher, V.~G. Kim, B.~Russell, and M.~Aubry.
\newblock {AtlasNet: A Papier-M\^ach\'e Approach to Learning 3D Surface
  Generation}.
\newblock In {\em CVPR}, 2018.

\bibitem{gwak2017weakly}
J.~Gwak, C.~B. Choy, M.~Chandraker, A.~Garg, and S.~Savarese.
\newblock Weakly supervised {3D} reconstruction with adversarial constraint.
\newblock In {\em 3DV}, 2017.

\bibitem{hspHane17}
C.~H{\"a}ne, S.~Tulsiani, and J.~Malik.
\newblock Hierarchical surface prediction for {3D} object reconstruction.
\newblock In {\em 3DV}. 2017.

\bibitem{kl2019capnet}
N.~K~L, P.~Mandikal, M.~Agarwal, and R.~V. Babu.
\newblock {CAPNet}: Continuous approximation projection for 3{D} point cloud
  reconstruction using 2{D} supervision.
\newblock In {\em AAAI}, 2019.

\bibitem{kar2017learning}
A.~Kar, C.~H{\"a}ne, and J.~Malik.
\newblock Learning a multi-view stereo machine.
\newblock In {\em NeurIPS}, pages 364--375, 2017.

\bibitem{kingma2014adam}
D.~P. Kingma and J.~Ba.
\newblock Adam: A method for stochastic optimization.
\newblock {\em arXiv preprint arXiv:1412.6980}, 2014.

\bibitem{lai2018fast}
W.-S. Lai, J.-B. Huang, N.~Ahuja, and M.-H. Yang.
\newblock Fast and accurate image super-resolution with deep laplacian pyramid
  networks.
\newblock {\em IEEE TPAMI}, 2018.

\bibitem{li2018pointcnn}
Y.~Li, R.~Bu, M.~Sun, and B.~Chen.
\newblock {PointCNN}.
\newblock {\em arXiv preprint arXiv:1801.07791}, 2018.

\bibitem{lin2018learning}
C.-H. Lin, C.~Kong, and S.~Lucey.
\newblock Learning efficient point cloud generation for dense {3D} object
  reconstruction.
\newblock In {\em AAAI}, 2018.

\bibitem{mandikal20183dlmnet}
P.~Mandikal, N.~K~L, M.~Agarwal, and R.~V. Babu.
\newblock {3D-LMNet}: Latent embedding matching for accurate and diverse 3{D}
  point cloud reconstruction from a single image.
\newblock In {\em Proceedings of the British Machine Vision Conference
  ({BMVC})}, 2018.

\bibitem{mandikal20183dpsrnet}
P.~Mandikal, N.~K~L, and R.~V. Babu.
\newblock {3D-PSRNet}: Part segmented 3{D} point cloud reconstruction from a
  single image.
\newblock In {\em 3D Reconstruction Meets Semantics Workshop ({ECCVW})}, 2018.

\bibitem{qi2017pointnet}
C.~R. Qi, H.~Su, K.~Mo, and L.~J. Guibas.
\newblock Pointnet: Deep learning on point sets for {3D} classification and
  segmentation.
\newblock {\em CVPR}, 2017.

\bibitem{qi2016volumetric}
C.~R. Qi, H.~Su, M.~Nie{\ss}ner, A.~Dai, M.~Yan, and L.~J. Guibas.
\newblock Volumetric and multi-view cnns for object classification on {3D}
  data.
\newblock In {\em CVPR}, 2016.

\bibitem{qi2017pointnet++}
C.~R. Qi, L.~Yi, H.~Su, and L.~J. Guibas.
\newblock Pointnet++: Deep hierarchical feature learning on point sets in a
  metric space.
\newblock In {\em NeurIPS}, pages 5105--5114, 2017.

\bibitem{Riegler2017THREEDV}
G.~Riegler, A.~O. Ulusoy, H.~Bischof, and A.~Geiger.
\newblock Octnetfusion: Learning depth fusion from data.
\newblock In {\em 3DV}, 2017.

\bibitem{Riegler2017CVPR}
G.~Riegler, A.~O. Ulusoy, and A.~Geiger.
\newblock Octnet: Learning deep {3D} representations at high resolutions.
\newblock In {\em CVPR}, 2017.

\bibitem{su2018splatnet}
H.~Su, V.~Jampani, D.~Sun, S.~Maji, V.~Kalogerakis, M.-H. Yang, and J.~Kautz.
\newblock Splatnet: Sparse lattice networks for point cloud processing.
\newblock In {\em CVPR}, 2018.

\bibitem{pix3d}
X.~Sun, J.~Wu, X.~Zhang, Z.~Zhang, C.~Zhang, T.~Xue, J.~B. Tenenbaum, and W.~T.
  Freeman.
\newblock Pix{3D}: Dataset and methods for single-image {3D} shape modeling.
\newblock In {\em CVPR}, 2018.

\bibitem{tatarchenko2017octree}
M.~Tatarchenko, A.~Dosovitskiy, and T.~Brox.
\newblock Octree generating networks: Efficient convolutional architectures for
  high-resolution {3D} outputs.
\newblock In {\em CVPR}, 2017.

\bibitem{tulsiani2017multi}
S.~Tulsiani, T.~Zhou, A.~A. Efros, and J.~Malik.
\newblock Multi-view supervision for single-view reconstruction via
  differentiable ray consistency.
\newblock In {\em CVPR}, volume~1, page~3, 2017.

\bibitem{ulusoy2015towards}
A.~O. Ulusoy, A.~Geiger, and M.~J. Black.
\newblock Towards probabilistic volumetric reconstruction using ray potentials.
\newblock In {\em 3DV}, 2015.

\bibitem{wang2017cnn}
P.-S. Wang, Y.~Liu, Y.-X. Guo, C.-Y. Sun, and X.~Tong.
\newblock O-cnn: Octree-based convolutional neural networks for {3D} shape
  analysis.
\newblock {\em ACM Transactions on Graphics (TOG)}, 2017.

\bibitem{wu2017marrnet}
J.~Wu, Y.~Wang, T.~Xue, X.~Sun, B.~Freeman, and J.~Tenenbaum.
\newblock Marrnet: {3D} shape reconstruction via 2.5 d sketches.
\newblock In {\em NeurIPS}, pages 540--550, 2017.

\bibitem{wu2016learning}
J.~Wu, C.~Zhang, T.~Xue, B.~Freeman, and J.~Tenenbaum.
\newblock Learning a probabilistic latent space of object shapes via {3D}
  generative-adversarial modeling.
\newblock In {\em NeurIPS}, pages 82--90, 2016.

\bibitem{wu20153d}
Z.~Wu, S.~Song, A.~Khosla, F.~Yu, L.~Zhang, X.~Tang, and J.~Xiao.
\newblock {3D} shapenets: A deep representation for volumetric shapes.
\newblock In {\em CVPR}, 2015.

\bibitem{yan2016perspective}
X.~Yan, J.~Yang, E.~Yumer, Y.~Guo, and H.~Lee.
\newblock Perspective transformer nets: Learning single-view {3D} object
  reconstruction without {3D} supervision.
\newblock In {\em NeurIPS}, pages 1696--1704, 2016.

\bibitem{yu2018pu}
L.~Yu, X.~Li, C.-W. Fu, D.~Cohen-Or, and P.-A. Heng.
\newblock Pu-net: Point cloud upsampling network.
\newblock In {\em CVPR}, 2018.

\bibitem{yuan2018pcn}
W.~Yuan, T.~Khot, D.~Held, C.~Mertz, and M.~Hebert.
\newblock Pcn: Point completion network.
\newblock In {\em 3D Vision (3DV), 2018 International Conference on}, 2018.

\bibitem{zhu2017rethinking}
R.~Zhu, H.~K. Galoogahi, C.~Wang, and S.~Lucey.
\newblock Rethinking reprojection: Closing the loop for pose-aware shape
  reconstruction from a single image.
\newblock In {\em ICCV}, 2017.

\end{thebibliography}
}

\end{document}